\documentclass[11pt]{article}

\PassOptionsToPackage{table}{xcolor}

\usepackage[final]{acl}

\usepackage{times}
\usepackage{latexsym}
\usepackage[T1]{fontenc}

\usepackage[utf8]{inputenc}

\usepackage{microtype}

\usepackage{inconsolata}

\usepackage{graphicx}
\usepackage{amsmath,hyperref}
\usepackage{amsfonts}
\usepackage{enumitem}
\usepackage{xcolor}
\usepackage{booktabs}
\usepackage{multirow}
\usepackage{stfloats}
\usepackage{makecell}
\usepackage{tabularx,booktabs}
\usepackage{subcaption}
\usepackage{algorithm}
\usepackage{algpseudocode}
\usepackage{amssymb}
\usepackage[most]{tcolorbox}
\newcolumntype{Y}{>{\ttfamily\footnotesize\raggedright\arraybackslash}X}
\newcolumntype{R}{>{\footnotesize\raggedright\arraybackslash}p{2.2cm}}

\newtcolorbox{promptbox}{
  colback=gray!10,
  colframe=gray!40,
  boxrule=0.3pt,
  arc=2pt,
  left=4pt, right=4pt, top=4pt, bottom=4pt
}

\title{SR-Fraud: An Outcome-Supervised Reflective LLM Agent Framework \\ for Non-Stationary Payment Fraud Detection}

\author{Xuwei Tan \thanks{Work done during internship at Coinbase. This paper contains the author’s personal opinions and does not constitute a company policy or statement. These opinions are not endorsed by or affiliated with Coinbase, Inc. or its subsidiaries.}  \\
  Coinbase, Inc.  \\
  The Ohio State University  \\
   \\\And
  Yao Ma \\
  Coinbase, Inc. \\
   \\
   \\\And
  Xueru Zhang \\
  The Ohio State University \\
   \\}

\begin{document}
\maketitle

\begin{abstract}
  Real-time payment fraud detection is a non-stationary streaming prediction
  problem: adversaries adapt before supervised labels mature, and localized burst
  attacks can cause losses before retraining. Production systems
  typically rely on tabular classifiers and rules, which can struggle to capture
these emerging sequential patterns before periodic retraining occurs. We present
\textbf{SR-Fraud}, an outcome-supervised reflective LLM framework that decouples
request-time decisions from offline adaptation. A frozen, stateless agent scores
each transaction from a Hybrid Episodic Window to track behavioral shifts,
while an offline reflection agent proposes boundary hypotheses from matured
errors. A deterministic verifier then admits only supported hypotheses into an
executable knowledge state. On a production payment-fraud benchmark, SR-Fraud
improves all detection metrics over its frozen decision agent, obtains higher
point estimates than static and periodically retrained CatBoost, and detects an
emerging fraud burst.
\end{abstract}

\section{Introduction}
\label{sec:intro}

Payment fraud in financial platforms is an inherently non-stationary streaming problem \citep{lu2018learning,dal2015credit}. As adversaries continuously adapt to bypass active defenses, fraud patterns learned in static training data quickly decay. This non-stationarity is particularly challenging during localized burst attacks, where adversaries execute high-value operations in a short window before a model can adapt to the new pattern.

Production pipelines typically rely on tabular classifiers like Gradient Boosted Decision Trees (GBDTs) \citep{chen2016xgboost, prokhorenkova2018catboost} and rule engines \citep{soui2019rule}. However, they face two limitations. First, fraudulent transactions are often individually plausible, becoming recognizable only when evaluated against historical user behavior. Standard GBDTs score rows independently unless temporal behavior is explicitly encoded through engineered aggregate features \citep{bahnsen2016feature}. Second, responding to distribution shifts requires periodic retraining and newly engineered features.

Large Language Models (LLMs) offer a complementary paradigm for non-stationary risk streams. Utilizing pre-trained world knowledge, LLMs naturally comprehend complex, high-cardinality metadata zero-shot (e.g., recognizing disposable email domains as potential risk signals), without manual feature engineering or model training. In addition, by comparing current transactions against a user's recent behavioral history via in-context learning \citep{brown2020fewshot}, LLMs intuitively uncover subtle anomalies and effectively handle transient fraud spikes. Finally, the explicit natural-language insights and rules synthesized by the LLM can be transferred across diverse product lines.

Despite these benefits, directly integrating generic LLM agents into inline fraud pipelines remains non-trivial due to two core bottlenecks. On one hand, real-time payment systems are governed by strict latency budgets. Advanced inference-time deliberation \citep{shinn2023reflexion, madaan2023selfrefine, yao2023tree}, while effective, induces unacceptable latency overhead, leading to a prominent \textbf{reasoning-budget gap}. On the other hand, deployed LLMs are frozen, meaning their general knowledge cannot natively adjust to dynamic data patterns, highly institution-specific fraud definitions, or operational policies. This causes a \textbf{adaptation gap} between pre-trained general capabilities and the need for fine-grained domain alignment.

To bridge these gaps, we introduce \textbf{SR-Fraud}, an outcome-supervised reflective LLM agent framework that decouples real-time inference from asynchronous offline learning. At request time, a stateless decision agent scores each transaction in a single pass over a \textbf{Hybrid Episodic Window}, combining recent user sequences with localized velocity signals. This ensures the model reasons over behavioral changes rather than isolated events, while staying within latency budgets. Offline, an \textbf{Outcome-Supervised Reflection Agent} analyzes matured errors\footnote{Maturation refers to the delay before true transaction labels are confirmed, typically via user-filed chargebacks.} to propose new rule-based hypotheses. Rather than relying on LLM self-critique, a deterministic harness verifies these proposals against historical labels. Verified rules accumulate into an \textbf{Evolving Expert Knowledge State}, a compact and human-readable set of rules that adjusts the decision agent's boundary cases. Our key contributions are summarized as follows:

\begin{itemize}[leftmargin=*,noitemsep,topsep=0.0pt]
    \item We introduce \textbf{SR-Fraud}, an architecture that keeps offline reflection off the request path and requires only \emph{one} decision-model invocation per reviewed transaction, bridging the \textit{reasoning-budget gap}  in production pipelines.
    \item We propose an \textbf{outcome-supervised reflection mechanism} that forms a compact symbolic knowledge state through propose-and-verify hypotheses on the decision agent's outcomes, closing the \textit{adaptation gap}.
    \item We benchmark existing LLM-based methods and SR-Fraud on \textbf{data from our production stream}, demonstrating the behavior of different LLMs and the benefits of SR-Fraud relative to traditional tabular classifiers.
    \item We further explore \textbf{fraud-domain LLM post-training} and analyze deployment trade-offs between token-priced commercial APIs and self-hosted open-source models.

\end{itemize}

\section{Methodology}
\label{sec:method}

\subsection{System Overview}

\begin{figure*}[t]
\centering
\includegraphics[width=1\textwidth]{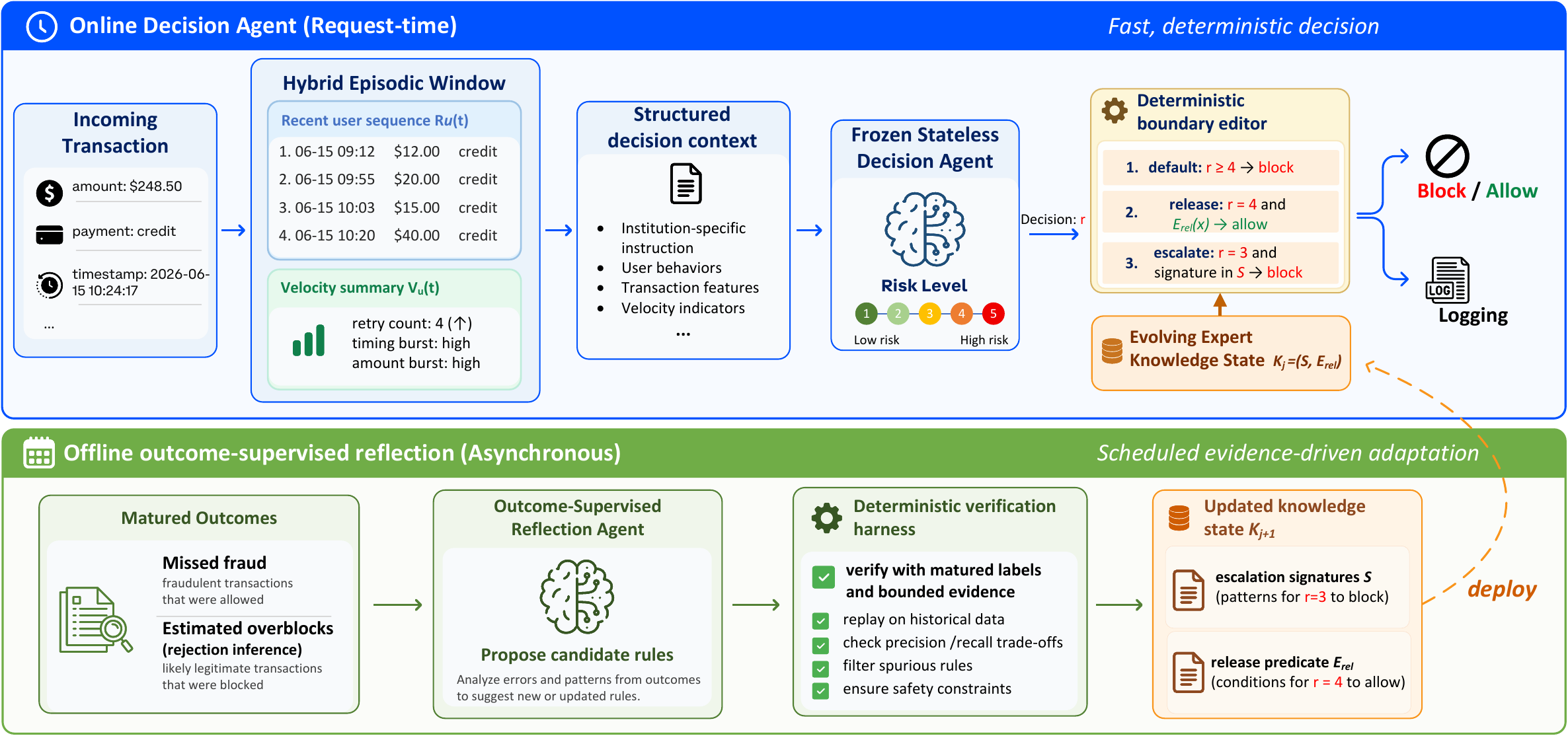}
\caption{
Overview of SR-Fraud. At request time, a fast stateless agent assigns a bounded
risk level from the current transaction and Hybrid Episodic Window, after which
the expert knowledge state deterministically adjusts boundary decisions.
Offline reflection uses matured missed fraud and estimated overblocks to update $K$ for the next
cycle.
}
\label{fig:method-overview}
\end{figure*}

In production deployments, risk systems process transactions through a standard machine learning model (e.g., GBDTs) with rule engines performing broad, low-latency screening, while SR-Fraud is selectively invoked on those high-risk or ambiguous edge cases requiring complex contextual reasoning. This modular design balances latency and compute costs while preserving the high-throughput guarantees of the primary risk pipeline. In this work, we focus specifically on the LLM agent side and omit the upstream model contents. Architecturally, SR-Fraud enforces a separation between request-time inference and asynchronous offline learning. A frozen, stateless \emph{decision agent} assigns each transaction one of five bounded risk levels from a Hybrid Episodic Window. The resulting boundary scores are then adjusted by an \emph{Evolving Expert Knowledge State}, a compact symbolic store of verified corrections maintained by the \emph{reflection agent}. Figure~\ref{fig:method-overview} shows the framework.

\subsection{Transaction-Grounded Decision Agent}
\label{sec:decision-agent}

At request time, SR-Fraud acts on a payment-risk stream. Given an incoming
transaction $x_t$ from user $u$, the decision agent must assign a risk level
before the payment settles. A transaction that appears plausible in isolation
may nevertheless depart sharply from the user's established behavior. We
therefore cast scoring as contextual anomaly detection rather than
single-record classification.

Let $R_u(t)=(x_{t_1},\dots,x_{t_m})$ be a bounded, time-ordered window of user
$u$'s $m$ transactions preceding time $t$. SR-Fraud formats this event stream
as a \emph{Hybrid Episodic Window},
\begin{equation}
W_u(t)=\bigl(R_u(t),V_u(t)\bigr),
\end{equation}
where $V_u(t)$ is a burst-and-velocity summary computed from $R_u(t)$. The
decision context is
\begin{equation}
c_t=\Phi\bigl(x_t,W_u(t)\bigr),
\end{equation}
where $\Phi$ serializes the evidence as structured text. The LLM
maps this context to a bounded risk level $r(x_t)\in\{1,\dots,5\}$. We take $r(x_t)\geq4$ as a block.

The Hybrid Episodic Window makes temporal comparison the agent's core
reasoning primitive. The sequence $R_u(t)$ retains operational attributes such
as payment rail, amount, and destination while excluding personally
identifiable information. The summary $V_u(t)$ complements the raw sequence
with retry, timing, amount, and channel-velocity signals. Presenting both views
preserves event-level context while making short bursts explicit.

The score is then further corrected by an \textit{Evolving Expert Knowledge State $K$}, which is maintained and updated by the \textit{reflection agent} (\S\ref{sec:reflection}). After the agent emits $r(x_t)$, $K$ edits only the boundary cases: \emph{escalating} an under-blocked $r{=}3$ transaction to a block, or \emph{releasing} an over-blocked $r{=}4$ transaction to an allow.
All adaptation therefore occurs through a deterministic boundary layer, leaving
the decision agent stateless and single-pass.

\subsection{Outcome-Supervised Reflection}
\label{sec:reflection}

Outcome-supervised reflection converts matured payment outcomes into actionable updates for the \emph{Evolving Expert Knowledge State}. Fraud labels become reliable only after real-world outcomes, such as chargebacks or investigations, mature. Once per weekly cycle, the reflection agent analyzes the available resolved errors and proposes candidate rules. A deterministic harness verifies these candidates against matured data. Thus, the LLM directs \emph{where to search}, while evidence determines \emph{what enters the active policy}.
We use $t$ for transaction time and $j$ for the weekly deployment cycle. Within cycle $j$, the state $K_j$ is fixed for all request-time decisions.

\paragraph{Boundary-Editing Expert Knowledge State:}
Reflection avoids rewriting the underlying LLM policy, mitigating the risk of overturning high-confidence decisions. Given the frozen agent's score $r(x) \in \{1,\dots,5\}$ where $r(x) \ge 4$ defaults to a block, the state $K$ edits only boundary cases: escalating suspicious $r(x)=3$ transactions and releasing over-blocked $r(x)=4$ transactions.

Suppressing the cycle index, write the active state as
$K=(\mathcal{S},E_{\mathrm{rel}})$, where $\mathcal{S}$ is a set of
conjunctive escalation signatures and $E_{\mathrm{rel}}$ is a Boolean release
predicate. We write $K_{\varnothing}=(\varnothing,\bot)$ for the empty state, where $\bot$ is the always-false release predicate. The final deterministic decision policy is below, where $[\cdot]$ denotes an
Iverson indicator:
\begin{equation}
\begin{aligned}
\operatorname{esc}(x) &= [r(x){=}3] \land {\textstyle \bigvee_{s\in\mathcal{S}}} s(x), \\
\operatorname{rel}(x) &= [r(x){=}4] \land E_{\mathrm{rel}}(x), \\
\operatorname{block}(x) &= \bigl([r(x){\geq}4] \lor \operatorname{esc}(x)\bigr) \land \neg\operatorname{rel}(x).
\end{aligned}
\end{equation}
Assuming a fixed-size release predicate, the boundary layer executes in $O(|\mathcal{S}|)$ time, and $K=K_{\varnothing}$ exactly reproduces the decision agent. Every $s\in\mathcal{S}$ is a conjunction over the leaf-feature vocabulary,
as is the release predicate $E_{\mathrm{rel}}$. The set $\mathcal{S}$ is
grown and pruned across cycles, whereas $E_{\mathrm{rel}}$ is a single release
condition.

\paragraph{Reflection Cycle and Selective Labels.}
To dynamically update this state $K$, the reflection agent analyzes the online policy's matured errors at each deployment cycle $j$. However, acquiring these errors faces a selective-labels asymmetry: missed fraud is easily observable from settled transactions, but false blocks censor their own counterfactual outcomes. Consequently, we compile two matured pools entirely from observable, allowed transactions from previous cycles: the missed-fraud pool ($\mathcal{F}_j$) to drive escalation mining, and the observable $r{=}3$ bank ($\mathcal{B}_{3,j}$) to ground release safety via \emph{rejection inference}. A legitimate-dense band in settled $\mathcal{B}_{3,j}$ traffic proxies the censored $r{=}4$ traffic, estimating the true positives a release would surrender without relying on unobserved block labels; we validate this proxy against an oracle in Appendix~\ref{app:rejection-validation}.

The harness renders this evidence into a textual digest. The offline LLM proposes
$\mathcal{G}_j=(\mathcal{A}_j,\mathcal{Q}_j,\mathcal{L}_j)$, where
$\mathcal{A}_j$ contains amount atoms, $\mathcal{Q}_j$ contains qualifier
atoms, and $\mathcal{L}_j$ contains candidate release bands. The deterministic harness enumerates the escalation language
$\mathcal{H}_j=\{a\land q_1[\land q_2]:
 a\in\mathcal{A}_j,\ q_1,q_2\in\mathcal{Q}_j\}$, where
$[\land q_2]$ denotes an optional second qualifier. It then verifies the
candidates and updates $K$. Thus, the LLM defines the search space, while the
harness controls validation (Algorithm~\ref{alg:reflection-cycle}).

\paragraph{Memory Evolution.}
Because deployed rules block matching transactions, their counterfactual
outcomes are no longer observed. We therefore re-evaluate all signatures each
cycle. To avoid pruning a pre-test-verified rule on a thin matured window, we
retire it only when its marginal-precision Wilson \emph{upper} bound is below the
operating threshold and it also fails the original train--validation gate.
Candidates that clear the gate are ranked by their Wilson lower bound and
admitted up to a fixed per-cycle cap. The knowledge state is updated once per
cycle from matured evidence; its initialization and replay trajectory are
detailed in Appendix~\ref{app:implementation-details}.

\paragraph{Verification by Bounded Evidence.}
To mitigate the winner's curse when evaluating many candidates on short time
windows, the harness uses one-sided Wilson confidence bounds. Suppressing the
cycle index $j$, let $p_{\mathrm{op}}$ be the decision agent's operating
precision, $p_{\mathrm{marg}}(h)$ the precision of transactions newly blocked
by escalation $h$, and $p_{\mathrm{free}}(b)$ the fraud rate among transactions
released by band $b$. We require:
\begin{equation}
\begin{aligned}
\operatorname{LCB}_{\mathrm{Wil}}(p_{\mathrm{marg}}(h))
&> \gamma\,p_{\mathrm{op}},\\
\operatorname{UCB}_{\mathrm{Wil}}(p_{\mathrm{free}}(b))
&< p_{\mathrm{op}}.
\end{aligned}
\end{equation}
Together with structural pre-tests, these bounds constrain boundary updates.

\section{Experiments}
\label{sec:experiments}

\subsection{Data and Replay Protocol}
\label{sec:data}

\textbf{Dataset.}
We evaluate a fixed cohort of $16{,}140$ transactions from nine weeks of a
production payment-risk stream with an overall $5.04\%$ fraud rate (Table~\ref{tab:benchmark});
it is a case--control sample (all matured fraud, subsampled legitimate traffic),
and every system is scored on all $16{,}140$ transactions. Training and
validation each cover two preceding months.\footnote{Company policy limits
disclosure of the broader stream's distribution and sampling statistics.}
Appendix~\ref{app:production-data} provides construction details.

\begin{table}[htbp]
\centering
\small
\begin{tabular}{lc}
\toprule
\textbf{Statistic} & \textbf{Benchmark} \\
\midrule
Transactions & 16{,}140 \\
Fraud transactions & 813 \\
Fraud rate & 5.04\% \\
Entities (P / N) & 7{,}293 (353 / 6{,}940) \\
Fraud per week & 48/86/73/38/73/67/90/\textbf{235}/103 \\
\bottomrule
\end{tabular}
\caption{Evaluation benchmark. P/N denotes entities with/without confirmed fraud; per-week fraud counts reveal a burst in week~8.}
\label{tab:benchmark}
\end{table}

\textbf{Replay protocol.}
We use time-ordered splits and out-of-time evaluation rather than random partitions. The validation period is used for model selection, and the final period is replayed in timestamp order. Earlier training and validation events may appear as prior transaction history, but their labels are never exposed to the decision agent.

\textbf{Metrics.}
We evaluate precision, recall, F1, and \textbf{dollar-weighted recall}, defined as the ratio of the total amount value of correctly blocked fraudulent transactions to the total value of all actual fraud.

\subsection{Implementation Details}
\label{sec:reflection-implementation}

We replay the test stream in weekly cycles with a fixed two-cycle maturation
lag: the policy for cycle $j$ may use outcomes only through cycle $j-2$. This
maturation wall is enforced by the replay harness, and every cycle is scored
once and then frozen, without retrospective re-scoring under a later knowledge
state. Before test replay, the same propose--verify pipeline is applied to the
preceding train--validation period; subsequent updates use only outcomes exposed
by the maturation wall. Full implementation details are provided in
Appendix~\ref{app:implementation-details}.

\subsection{Comparison Systems and Baselines}
\label{sec:baselines}

We compare systems that isolate tabular learning, the LLM prior, behavioral
history, and outcome-supervised adaptation.

\begin{itemize}[leftmargin=*,noitemsep,topsep=0.0pt]
    \item \textbf{CatBoost.} It is trained on the full training period, with hyperparameter optimization using Optuna. We also evaluate \textbf{CatBoost-periodical}, the same model periodically retrained over the stream.
    \item \textbf{Seq-GRU.} A supervised GRU over the last-$k$ transaction sequence as the decision agent \citep{jurgovsky2018sequence}.
    \item \textbf{Direct LLM. } The LLM scores only the current transaction \citep{hegselmann2023tabllm}.
    \item \textbf{Decision Agent-only.} The SR-Fraud decision agent without reflection or an evolving knowledge state.
    \item \textbf{Reflexion.} A weekly verbal-memory baseline adapted from
    \citet{shinn2023reflexion}.
    \item \textbf{SR-Fraud.} The decision agent augmented with our outcome-supervised reflection mechanism.
\end{itemize}
Full feature, training, and adaptation details for all baselines are provided in
Appendix~\ref{app:baseline-details}.

\subsection{Results and Analysis}
\label{sec:analysis}

\subsubsection{Overall Detection Performance}
\label{sec:overall}

\begin{table*}[t]
\centering
\small
\begin{tabular}{lccccccc}
\toprule
\textbf{Method} & \textbf{$\Delta$Prec.} & \textbf{$\Delta$Recall} &
\textbf{$\Delta$F1} & \textbf{$\Delta$Recall$_{\$}$} & \textbf{Block} &
\textbf{$p_{90}$ (s)} & \textbf{Mean (s)} \\
\midrule
\quad CatB. (ref.) & 0.0 & 0.0 & 0.0 & 0.0 & 5.00\% & $<$0.1 & $<$0.1 \\
\quad CatBoost-periodical & +4.2 & +4.1 & +4.2 & +3.9 & 5.00\% & $<$0.1 & $<$0.1 \\
\quad Seq-GRU & -2.5 & +6.5 & +1.4 & +4.0 & 5.00\% & $<$0.1 & $<$0.1 \\
\midrule
\multicolumn{8}{l}{\emph{Direct LLM} (current transaction only)}\\
\quad GPT-5.4 & -18.5 & -4.0 & -13.7 & -15.5 & 10.32\% & 1.06 & 0.81 \\
\quad Opus~4.5 & -19.2 & -7.1 & -15.0 & -18.0 & 9.68\% & 2.01 & 1.35 \\
\quad Sonnet~4.6 & -8.0 & -21.6 & -17.5 & -35.7 & 2.16\% & 1.71 & 1.21 \\
\midrule
\multicolumn{8}{l}{\emph{Decision Agent-only}: SR-Fraud online agent, no reflection}\\
\quad GPT-5.4 & -11.8 & +2.8 & -6.4 & -11.0 & 8.60\% & 1.70 & 1.27 \\
\quad Opus~4.5 & -2.1 & -1.5 & -1.8 & -15.3 & 5.10\% & 1.97 & 1.28 \\
\quad Sonnet~4.6 & -0.9 & +5.0 & +1.8 & -7.6 & 5.97\% & 2.02 & 1.44 \\
\midrule
\multicolumn{8}{l}{\emph{Reflexion}: verbal reflection (risk~$\geq\!4$)}\\
\quad GPT-5.4 & -25.1 & +25.4 & -19.6 & +10.8 & 41.44\% & 2.48 & 2.22 \\
\quad Opus~4.5 & -22.2 & +31.7 & -14.8 & +19.5 & 32.24\% & 2.81 & 2.00 \\
\quad Sonnet~4.6 & -13.3 & +23.5 & -3.9 & +26.1 & 14.80\% & 2.85 & 1.70 \\
\midrule
\multicolumn{8}{l}{\emph{SR-Fraud} (full system)}\\
\quad GPT-5.4 & -1.2 & +7.6 & +2.7 & +2.8 & 6.44\% & 1.71 & 1.28 \\
\quad Opus~4.5 & \textbf{+6.2} & +3.0 & +4.6 & -2.2 & 4.59\% & 1.97 & 1.28 \\
\quad Sonnet~4.6 & +4.5 & \textbf{+12.3} & \textbf{+8.0} & \textbf{+11.7} & 6.08\% & 2.03 & 1.45 \\
\bottomrule
\end{tabular}
\caption{Overall performance as percentage-point differences from static CatBoost. Block rates and latency remain absolute.}
\label{tab:main-results}
\end{table*}

Table~\ref{tab:main-results} yields three main findings.

\textbf{Fraud is hard to discover from a single transaction, yet LLMs are
strong at reasoning over behavior.} Relative to static CatBoost, the direct
Sonnet~4.6 reviewer has deltas of $-8.0$ precision, $-21.6$ recall, $-17.5$ F1,
and $-35.7$ dollar-weighted recall. Adding recent history changes these gaps to
$-0.9$, $+5.0$, $+1.8$, and $-7.6$, respectively. Opus and GPT-5.4 show the
same direction, indicating that behavioral comparison supplies the main signal.

\textbf{Reflection improves precision, recall, and dollar-weighted recall
together.} Relative to the Sonnet~4.6 decision agent, full SR-Fraud improves
precision by $+5.4$, recall by $+7.3$, F1 by $+6.2$, and dollar-weighted recall
by $+19.3$ points, at essentially the same block rate
($5.97\%\!\to\!6.08\%$). All four paired bootstrap intervals exclude zero,
including $+6.2$ F1 $[3.6,9.0]$ and $+19.3$ Recall$_{\$}$ $[14.3,24.5]$.
Against CatBoost-periodical at its $5\%$ point, the deltas are $+0.3$
precision, $+8.1$ recall, $+3.8$ F1, and $+7.8$ dollar recall; recall and
dollar recall exclude zero, while precision and F1 do not. On cohorts with at
least one prior transaction, the recall, F1, and dollar-recall advantages over
CatBoost-periodical are significant; relative to the decision agent, the
dollar-recall gain remains significant in the $\geq1$ and $\geq2$ cohorts
(Appendix~\ref{app:bootstrap}).

The larger dollar-recall gain is consistent with feedback asymmetry: missed
fraud provides direct labels for escalation, whereas releases rely on a
conservative proxy for censored blocks. This favors recovery of value-skewed
missed fraud.

\textbf{Free-text reflection can destabilize the operating point.}
Reflexion results show that purely verbal reflection, which summarizes past
outcomes in a free-text memory, can accumulate overly broad or aggressive
patterns and cause the LLM to block more legitimate transactions. SR-Fraud
instead converts reflection into structured candidate rules and verifies each
rule against matured evidence before admission. This shows that unverified
free-text reflection can be unstable at the required operating point.

\subsubsection{Behavioral Context Drives LLM}
\label{sec:context-depth}

Table~\ref{tab:cohort-perf} evaluates fixed models across cumulative history
cohorts. As history deepens, the LLM advantage over both CatBoost variants widens with history depth.
Although CatBoost includes rolling
history aggregates, these fixed summaries may encode a broad prior that entities
with longer histories are safer, while the LLM instead compares the current event directly with the raw episodic sequence.
The widening gap motivates further routing history-rich traffic to the
contextual LLM layer.

\begin{table*}[t]
\centering
\small
\setlength{\tabcolsep}{4pt}
\begin{tabular}{lr rrrr rrrr}
\toprule
 & & \multicolumn{4}{c}{\textbf{$\Delta$Precision}} &
 \multicolumn{4}{c}{\textbf{$\Delta$Dollar-recall}} \\
\cmidrule(lr){3-6}\cmidrule(lr){7-10}
\textbf{Prior history} & \textbf{$n$} & CatB. & \makecell{CatB.\\period.} &
\makecell{Dec.\\agent} & SR-Fraud & CatB. & \makecell{CatB.\\period.} &
\makecell{Dec.\\agent} & SR-Fraud \\
\midrule
$\geq0$ (all) & 16{,}140 & 0.0 & +4.2 & -0.9 & \textbf{+4.5} & 0.0 & +3.9 & -7.6 & \textbf{+11.7} \\
$\geq1$ prior & 10{,}263 & 0.0 & +5.8 & +7.5 & \textbf{+9.5} & 0.0 & +7.9 & +9.9 & \textbf{+22.3} \\
$\geq2$ prior & 7{,}391 & 0.0 & +8.6 & +15.0 & \textbf{+15.0} & 0.0 & +11.4 & +27.1 & \textbf{+35.0} \\
$\geq3$ prior & 5{,}650 & 0.0 & +11.9 & +21.2 & \textbf{+21.8} & 0.0 & +13.6 & +33.1 & \textbf{+43.5} \\
\bottomrule
\end{tabular}
\caption{Performance by prior-history depth as percentage-point differences from static CatBoost within each cohort.}
\label{tab:cohort-perf}
\end{table*}

\subsubsection{Detection of an Emerging Fraud Burst}
\label{sec:burst}

We isolate the largest test-stream burst, a coordinated low-value
fraud attack comprising $206$ of the $235$ confirmed frauds in
week~8, and separately examine a predefined high-value fraud subset
($55$ frauds) where prevented loss is concentrated (Table~\ref{tab:burst}). The
subset definition was fixed before evaluating model performance; its numerical
cutoff is withheld under internal disclosure policy.

\begin{table}[t]
\centering
\footnotesize
\setlength{\tabcolsep}{2.5pt}
\begin{tabular}{lcc}
\toprule
\textbf{Detector} & \textbf{$\Delta$Recall} & \textbf{$\Delta$Recall$_{\$}$} \\
\midrule
\multicolumn{3}{l}{\emph{Week-8 burst} ($n=206$)}\\
\quad CatBoost (reference) & 0.0 & 0.0 \\
\quad CatBoost-periodical & +9.2 & +6.6 \\
\quad Decision agent (Sonnet~4.6) & +38.4 & +34.8 \\
\quad SR-Fraud & \textbf{+41.7} & \textbf{+46.4} \\
\midrule
\multicolumn{3}{l}{\emph{Emerging high-value vector} ($n=55$)}\\
\quad CatBoost (reference) & 0.0 & 0.0 \\
\quad CatBoost-periodical & -3.7 & -1.9 \\
\quad Decision agent (Sonnet~4.6) & -16.4 & -15.3 \\
\quad SR-Fraud & \textbf{+10.9} & \textbf{+10.4} \\
\bottomrule
\end{tabular}
\caption{Burst and high-value performance as percentage-point differences from static CatBoost within each subset. Other systems retain their native operating points.}
\label{tab:burst}
\end{table}

\textbf{The base LLM detects the unseen burst zero-shot.} Both CatBoost and CatBoost-periodical miss much of the burst because their latest
models predate the emerging pattern. At their native operating points, the LLM identifies the burst
from recent transaction context without first observing labeled examples of the
attack; Appendix~\ref{app:reverse-budget-match} confirms the comparison at an exact $6\%$ budget.

\textbf{Reflection complements burst detection with high-value coverage.}
On the burst, SR-Fraud exceeds static CatBoost at its native $5\%$ point by $+41.7$
recall points and $+46.4$ dollar-recall points, compared with $+38.4$ and
$+34.8$ for the decision agent. On the high-value subset, SR-Fraud's deltas are
$+10.9$ and $+10.4$, whereas the decision agent's are negative.
Under the two-cycle maturation wall, week-8
outcomes become available only after the replay ends, so reflection cannot have
learned from this burst itself. The two components therefore play distinct
roles: the base LLM detects the new volume burst from behavioral context, while
previously verified boundary rules found on earlier matured cycles improve
high-value coverage.

\subsubsection{Post-Training the Decision Agent}
\label{sec:post-training}

While the central contribution of this work is the reflection
framework, we additionally explore post-training the decision agent itself. We
include this study because its performance, cost, and latency implications are of
interest to industry practitioners deploying LLM risk agents at scale;
concretely, we compare a commercial API backbone against a self-hosted,
domain-adapted model along these three axes.

On cost and latency, commercial APIs reduce infrastructure work but introduce
per-token pricing, variable network latency, and limited control over model
upgrades; and on performance, newer Opus, Sonnet, and GPT releases in fact regress
on this benchmark (Table~\ref{tab:model-generation}). At sustained traffic volume,
self-hosting instead trades per-request charges for predictable GPU capacity while
enabling domain-specific training and tighter data control. 
We therefore post-train a self-hosted Qwen3.5-9B decision agent
\citep{qwen35} using outcome-supervised rewards with
the SkyRL framework \citep{griggs2025skrylv01}. Despite its
smaller size and lack of a symbolic knowledge state, SR-Fraud-PT attains similar detection
performance while reducing
request latency and, at sufficient sustained volume, modeled serving cost.
Table~\ref{tab:post-training-capacity} reports the result;
Appendix~\ref{app:post-training-details} gives the training recipe and latency
measurement.

\begin{table}[t]
\centering
\footnotesize
\setlength{\tabcolsep}{2pt}
\begin{tabular}{lccccc}
\toprule
\textbf{System} & \textbf{$\Delta$P} & \textbf{$\Delta$R} &
\textbf{$\Delta$F1} & \textbf{$\Delta$R$_{\$}$} & \textbf{$p_{90}$ (s)} \\
\midrule
CatBoost (reference) & 0.0 & 0.0 & 0.0 & 0.0 & $<$0.1 \\
Decision Agent & -0.9 & +5.0 & +1.8 & -7.6 & 2.02 \\
SR-Fraud & +4.5 & +12.3 & +8.0 & +11.7 & 2.03 \\
\midrule
Base 9B (no post-train) & -18.8 & -9.1 & -15.2 & -25.0 & 0.94 \\
\textbf{SR-Fraud-PT (9B)} & +5.4 & +10.5 &
+7.8 & +20.1 & 0.94 \\
\bottomrule
\end{tabular}
\caption{Commercial and self-hosted systems as percentage-point differences
from static CatBoost. Latency remains absolute.}
\label{tab:post-training-capacity}
\end{table}

\section{Conclusion}
\label{sec:conclusion}

We presented \textbf{SR-Fraud}, which combines a stateless contextual LLM with
outcome-supervised symbolic reflection for non-stationary payment fraud
detection. A deterministic verifier admits and retires boundary rules from
matured evidence, preserving one request-time LLM call and an auditable
knowledge state. Chronological replay shows that behavioral context and verified
reflection improve fraud capture over static and periodically retrained tree
baselines; an exact block-count comparison preserves the advantage while
holding customer friction fixed.

\section*{Limitations}

\paragraph{Confidence bounds.}
The verifier uses a heuristic one-sided Wilson bound without a formal
multiple-testing correction. Re-verifying the persisted main-run candidates
across the tested confidence settings yields a stable operating point
(Appendix~\ref{app:z-sensitivity}); this sensitivity check does not, however,
constitute formal control of adaptive multiple comparisons.

\paragraph{Hosted-model stochasticity.}
Performance is stable across the evaluated reflection/decision-backbone
configurations. Repeated full-stream decision replays were not conducted because
of their API and computational cost. The entity bootstrap quantifies sampling
uncertainty.

\paragraph{Proprietary evaluation and generality.}
The study uses one non-releasable production stream. Public fraud benchmarks are
typically synthetic or highly anonymized and do not preserve the metadata,
entity histories, and delayed-label dynamics required here. The
production-derived cohort improves realism but limits independent replication
and cross-platform generalization. Company policy also precludes disclosure of
the legitimate sampling rate, upstream routing logic, represented traffic share,
and production prevalence; we therefore make no end-to-end production-rate claim.
Metrics are reported as percentage-point differences relative to the corresponding 
CatBoost reference.

\paragraph{Fixed maturation lag.}
The two-cycle delay approximates the dominant label-maturation pattern summarized
from the historical production stream. The replay uses this fixed wall rather
than transaction-specific availability times, providing a reproducible
approximation while abstracting away proprietary timing details.

\paragraph{Hosted-provider latency.}
The reported hosted percentiles are point-in-time, request-level measurements
through model providers for this benchmark, rather than production SLO guarantees
or full application latency. Provider-side load, request routing, service changes,
and network conditions can change these values.

\paragraph{Off-policy replay.}
The replay applies SR-Fraud's decisions without altering the stream, so a
transaction the policy would have blocked still appears in later histories.
Confirmed fraud usually triggers entity denylisting, making subsequent
same-entity fraud rare after confirmation. This reduces the issue: pre-confirmation attempts and legitimate counterfactual blocks could
still reshape histories. Policy-consistent or shadow evaluation is left to
future work.

\section*{Ethics Statement}

The production system operates on payment transaction data.
Direct identifiers and designated PII fields are excluded from model prompts.
Our reported experiments run the reflection loop fully automatically; in
production deployment, disputes are reviewed by a human analyst as part of normal
operations and the reflection-loop knowledge update passes an optional
human-review checkpoint before deployment.

Only a vetted subset of transaction attributes enters the prompt, and content is
serialized as data rather than instructions. These controls reduce, but do not
eliminate, metadata, re-identification, or prompt-injection risk.

\bibliography{custom}

\clearpage
\appendix

\section{Related Work}
\label{sec:related}

\paragraph{Payment fraud detection.}
Payment fraud detection operates on heterogeneous transaction data under
evolving behavior and delayed supervision. Prior work compares data-mining
classifiers \citep{bhattacharyya2011data}, studies temporal feature engineering
\citep{bahnsen2016feature}, and analyzes concept-drift adaptation generally
\citep{gama2014survey} and with delayed fraud labels specifically
\citep{dal2015credit}. Sequence-based models directly exploit ordered card
activity \citep{jurgovsky2018sequence}, while SCARFF studies scalable streaming
credit-card fraud detection \citep{carcillo2018scarff}. TitAnt presents an
online, real-time transaction-fraud system at Ant Financial
\citep{cao2019titant}. Scalable tree learners such as XGBoost, LightGBM, and
CatBoost provide efficient tabular baselines
\citep{chen2016xgboost,ke2017lightgbm,prokhorenkova2018catboost}. SR-Fraud is
complementary to this literature: it retains the tabular first pass but adds
contextual LLM reasoning and a verified symbolic adaptation layer.

\paragraph{LLMs and foundation models for financial and tabular data.}
TabLLM studies few-shot classification by representing tabular examples as text
\citep{hegselmann2023tabllm}, while UniPredict investigates LLMs as general
tabular classifiers \citep{wang2023unipredict}. More recently, TLRD
\citep{liang2026tlrd} distills instance-, dataset-, and
comparison-level rationales into compact LLMs for tabular
prediction and explanation. TabPFN provides a complementary
foundation-model approach specialized for small tabular classification problems
\citep{hollmann2023tabpfn}. In finance, FinBERT targets financial text
\citep{yang2020finbert}, and BloombergGPT studies domain-specific language-model
pretraining for finance \citep{wu2023bloomberggpt}. In fraud-specific settings,
\citet{tan2025understanding} present a case study of LLMs on structured financial
data, and \citet{qu2025llm} study LLM-enhanced self-evolving reinforcement
learning for multi-step e-commerce fraud detection. SR-Fraud instead
focuses on a frozen request-time scorer and outcome-supervised symbolic updates,
without online parameter optimization.

\paragraph{Reflective and self-improving agents.}
Prior agents improve behavior through verbal memory or iterative feedback:
Reflexion stores task-derived reflections \citep{shinn2023reflexion}, Self-Refine
revises outputs using self-feedback \citep{madaan2023selfrefine}, and Tree of
Thoughts searches intermediate reasoning states \citep{yao2023tree}. Related
systems interleave reasoning and actions \citep{yao2023react}, combine persistent
memory with reflection \citep{park2023generative}, use external-tool feedback
\citep{gou2024critic}, or request targeted human guidance to update a knowledge
repository \citep{he2025enabling}. SR-Fraud instead proposes hypotheses from
weekly error batches and admits a boundary rule only after deterministic
verification against matured labels; its replay requires no human guidance.

\section{SR-Fraud Implementation Details}
\label{app:implementation-details}

\subsection{Verification and Replay Configuration}

\paragraph{Reflection Configuration.}
All reflection constants are fixed across cycles.
The pre-test rarity ceiling is $\rho=0.020$ of legitimate traffic. Its
conditional-precision floor is $0.30$ on train--validation, with at least 10
matched frauds. Hyperparameters are chosen based on the validation phase.
Both one-sided Wilson bounds use $z=1.0$, and the escalation admission margin is
$\gamma=1.15$. An escalation requires at least 8 marginal rows and 2 new true
positives. For per-cycle corroboration, a cycle votes after 2 matched boundary
rows, and a candidate needs at least 2 voting cycles. Retirement requires at
least 8 marginal rows, a marginal Wilson upper bound below the operating floor,
and failure of the original train--validation gate. A release band requires 20 settled boundary rows, permits at most
one dissenting cycle, and must be exercised by settled data to at least half its
proposed ceiling. Up to five escalation signatures that clear the gate may be
admitted per cycle, ordered by their one-sided Wilson lower bound on marginal
precision.

\paragraph{Temporal protocol.}
Cycles are weekly with a two-cycle maturation lag. To model
production label delay conservatively and reproducibly, the replay does not
reveal a cycle's outcomes until two later cycles, even though the offline
benchmark physically contains every matured label. Reflection runs once per
cycle as an offline agentic workflow and may use multiple model calls, none of
which lies on the request path. The policy for cycle $j$ is formed only from
cycles at most $j-2$. Each cycle is scored once and
frozen; aggregate results never re-score an earlier cycle under a later $K$.
The empty state $K_{\varnothing}$ exactly reproduces the frozen decision agent.
Although the backtest records physically contain settled labels for every
cycle, the digest builder exposes only outcomes permitted by the maturation
wall; this restriction is enforced by the harness rather than by agent
behavior.

\paragraph{Pre-replay construction.}
Before test replay, SR-Fraud constructs an initial knowledge state by applying
the same propose--verify procedure to the training and validation splits. The
anonymized atom vocabulary is generated at this point. During
replay, the weekly loop updates the deployed state using only matured
test-stream outcomes. The complete replay trajectory is retained in the internal
audit artifacts.

Algorithm~\ref{alg:reflection-cycle} summarizes the automated update applied
before cycle $j$ (\S\ref{sec:reflection}). Our experiments deploy $K_j$
directly; in production, an optional human-review checkpoint may gate its
deployment (\S\ref{sec:discussion}).

\begin{algorithm}[t]
\small
\caption{Automated reflection and deployment for cycle $j$.}
\label{alg:reflection-cycle}
\begin{algorithmic}[1]
\Require Matured data $\mathcal{D}_{\leq j-2}$; prior state
$K_{j-1}=(\mathcal{S}_{j-1},E_{\mathrm{rel},j-1})$
\State $d_j\gets\Call{Digest}{\mathcal{D}_{\leq j-2},K_{j-1}}$
\State $\mathcal{G}_j\gets\Call{ReflectAgent}{d_j}$, where
$\mathcal{G}_j=(\mathcal{A}_j,\mathcal{Q}_j,\mathcal{L}_j)$
\State $\mathcal{H}_j\gets\Call{Enumerate}{\mathcal{A}_j,\mathcal{Q}_j}$
\State $(\mathcal{V}^{\mathrm{esc}}_j,\mathcal{V}^{\mathrm{rel}}_j)
\gets\Call{Verify}{\mathcal{H}_j,\mathcal{L}_j,
\mathcal{D}_{\leq j-2},K_{j-1}}$
\State $E_{\mathrm{rel},j}\gets\Call{SelectRelease}{\mathcal{V}^{\mathrm{rel}}_j,
E_{\mathrm{rel},j-1}}$
\State $\widetilde{\mathcal{S}}_j\gets\Call{Retire}{\mathcal{S}_{j-1},
\mathcal{D}_{\leq j-2},E_{\mathrm{rel},j}}$
\State $\mathcal{S}_j\gets\Call{AdmitBest}{\widetilde{\mathcal{S}}_j,
\mathcal{V}^{\mathrm{esc}}_j}$
\State $K_j\gets(\mathcal{S}_j,E_{\mathrm{rel},j})$
\State $\Call{DecideCycle}{j,K_j}$
\Ensure Updated state $K_j$
\end{algorithmic}
\end{algorithm}

\subsection{Velocity Summary Computation}
\label{app:velocity}

The velocity block is computed over the recent transaction history as follows.
Timestamps are parsed to UTC datetimes;
inter-arrival intervals are computed between consecutive transactions
and summarised as minimum and median (in minutes).
Counts in tail windows (last 10 and 30 minutes before the current transaction)
are computed by scanning the sorted timestamp list.
Unique-set diversities of multiple important fields
are computed from the respective raw fields after stripping null/NaN values.
Historical amounts are extracted and summarized.

\subsection{Decision Agent Prompt}
\label{app:system-prompt}

\begin{promptbox}
\textbf{Decision Agent System Prompt} {\footnotesize(abridged; platform and
proprietary feature identifiers anonymized)}\par\medskip
{\footnotesize
\textbf{Role.} You are a fraud-detection assistant at [Platform], reviewing a
transaction. The user has $m$
prior successful transactions, but recency does not guarantee trust. A
compromised account or a card-testing burst can produce a deceptively consistent
history. Decide ALLOW or
BLOCK; ``fraud'' means the user is likely to dispute the transaction (financial
loss).\par\smallskip
\textbf{Risk scale.} 5 Critical (block); 4 High (block/verify); 3 Medium (add friction); 2 Low (allow+monitor); 1 Very Low (allow).\par\smallskip
\textbf{Data sources.} transaction/checkout fields;
[provider's signals]; [provider's signals] (feature names
anonymized).\par\smallskip
\textbf{History.} up to 15 most recent transactions (oldest$\rightarrow$newest,
each annotated with elapsed time) plus a computed burst/velocity block; treated as
context, not automatic evidence of legitimacy.\par\smallskip
\textbf{Output.} exactly one line: \texttt{RISK\_LEVEL: <1-5>}, with no analysis
or explanation.}
\end{promptbox}
The parser takes the first \texttt{RISK\_LEVEL:\textbackslash s*([1-5])} match;
a response with no match is treated as \emph{allow} (never a block), so a
malformed generation cannot silently raise the block rate.

\subsection{Reflection Agent Prompt}
\label{app:reflection-prompt}

\begin{promptbox}
\textbf{Reflection Agent Instructions} {\footnotesize(abridged schema;
leaf-feature values anonymized)}\par\medskip
{\footnotesize
\textbf{Role.} During each offline cycle, construct a structured hypothesis
space indicating \emph{where to look} for boundary errors. The agent may refine
its proposal over multiple model calls, but only the deterministic verifier can
admit or retire a rule.\par\smallskip
\textbf{Evidence digest (input).} matured per-cycle operating statistics; the
composition of the risk-3 bank; confirmed fraud that remains allowed under the
current knowledge state; and fraud-vs-legitimate profiles for eligible
features.\par\smallskip
\textbf{Final output (strict JSON).}
\texttt{\{"amount\_grid":[\dots], "qualifiers":[\dots],}
\texttt{"release\_grid":[\dots], "rationale":"\dots"\}}.
The \texttt{rationale} is retained for audit but never executed.
\texttt{amount\_grid}: permitted numeric atoms;
\texttt{qualifiers}: up to ${\sim}30$ single leaf-feature atoms;
\texttt{release\_grid}: a few conjunctive release bands.\par\smallskip
\textbf{Atom grammar.} Single atoms are numeric
\texttt{FIELD OP VALUE} ($\ge,\le,>,<,=$), categorical equalities over the
closed leaf-feature set, or named Boolean flags. Only
\texttt{release\_grid} may join permitted atoms with \texttt{\&}; the harness
constructs escalation conjunctions from \texttt{amount\_grid} and
\texttt{qualifiers}. Any atom outside the fixed vocabulary is dropped.\par\smallskip
\textbf{Constraints.} Propose candidate feature families and ranges, not
deployable decisions; prefer general leaf-level qualifiers; and do not bind proposals to
the decision agent's emitted risk level.}
\end{promptbox}

\section{Experimental Setup and Baseline Details}
\label{app:experimental-details}

\subsection{Benchmark Construction}
\label{app:production-data}

The production payment-risk stream is proprietary and cannot be released. We
 characterize the fixed evaluation cohort used for all reported
results \ref{tab:cohorts}. It contains $16{,}140$ upstream-allowed transactions with matured
outcomes, including 813 confirmed fraud cases ($5.04\%$ of the benchmark).
Company policy prevents disclosure of the broader-stream distribution and
sampling statistics. Direct identifiers and label-leaking fields are removed
before serialization, and point-in-time prior history is preserved for every
retained transaction.

The cohort is a case--control sample: all matured fraud is retained and
legitimate traffic is subsampled, so $5.04\%$ is not the production prevalence.
All systems score the same $16{,}140$ transactions. Recall is invariant to this
negative subsampling, whereas precision, F1, and block rate describe the
benchmark operating point. Production prevalence, routing criteria, and traffic
coverage remain proprietary. We therefore interpret results conditionally on
this cohort rather than as end-to-end production precision or block rates.

For chronological replay the benchmark is ordered by timestamp and partitioned
into nine weekly cycles (\S\ref{sec:reflection-implementation}). Model training
and operating-point selection for all systems use earlier, disjoint periods of
the same stream that are not part of the evaluation cohort. All splits are
disjoint at the transaction level, with no row-level overlap by transaction ID
or by the pair of entity hash and timestamp; recurring entities may appear
across periods, as expected in a streaming payment setting, and are used only to
construct prior-history features from events preceding the current transaction.

\paragraph{Data availability.}
Legal and compliance restrictions prevent release of transaction-level records.
To support scrutiny without exposing customer data, we report aggregate cohort
statistics and provide the replay protocol, prompts, model settings, and
verification procedure in the paper and appendix.

\begin{table}[h]
\centering
\small
\begin{tabular}{lccc}
\toprule
\textbf{Prior history} & \textbf{Transactions} & \textbf{Fraud} & \textbf{Fraud rate} \\
\midrule
$\geq 0$ (all) & 16{,}140 & 813 & 5.04\% \\
$\geq 1$ prior & 10{,}263 & 516 & 5.03\% \\
$\geq 2$ prior &  7{,}391 & 346 & 4.68\% \\
$\geq 3$ prior &  5{,}650 & 223 & 3.95\% \\
\bottomrule
\end{tabular}
\caption{Evaluation-cohort composition by available prior-history depth.}
\label{tab:cohorts}
\end{table}

\subsection{Baseline Implementations}
\label{app:baseline-details}

\subsubsection{Supervised CatBoost}

We train CatBoost on an earlier period using more than 100 features, including
rolling user-history aggregates. Identifiers and post-outcome fields are
excluded, while categorical variables are consumed natively. We use a positive-
class weight and early stopping on the subsequent validation period.

We run 50 hyperparameter-optimization trials with Optuna. The search space covers
the principal CatBoost capacity, step-size, sampling, and regularization
parameters, including tree depth, learning rate, L2 leaf regularization, random
strength, bagging temperature, and feature subsampling. We select the trial
with the highest validation PR-AUC. For the operating-point comparison, we evaluate the continuous CatBoost score
at a $5.0\%$ block-rate operating point. Table~\ref{tab:main-results} reports
this operating point alongside the realized block rates of the discrete LLM
actions.

To reflect production adaptation, \textbf{CatBoost-periodical} retrains the same
model monthly over the evaluation stream. Retraining respects the same
two-cycle maturation lag as SR-Fraud, and the validation split is used only for
early stopping.

\subsubsection{Supervised Sequence Model (Seq-GRU)}

\textbf{Seq-GRU} is a matched-input supervised baseline that controls whether the
gain comes from access to episodic history or from reasoning over it. It receives
the current transaction and up to $k{=}10$ strictly earlier events from the same
history index, using the CatBoost feature set with rolling aggregates disabled. A
per-event ReLU--dropout projection feeds a three-layer GRU (hidden size $128$,
packed variable-length sequences) and binary classification head. We train with
class-weighted binary cross-entropy and Adam, selecting hyperparameters and
early-stopping on validation average precision. Seq-GRU emits a continuous score, so as with CatBoost we evaluate it at a 5.0\% block-rate operating point on the evaluation cohort.

\subsubsection{LLM Comparison Systems}

The \textbf{Direct LLM} receives only the current transaction serialized as
text \citep{hegselmann2023tabllm}; it is the no-history control for the model's
single-transaction prior. The \textbf{Decision Agent-only} uses the same online
decision procedure but additionally receives the Hybrid Episodic Window
$W_u(t)$, enabling comparison against the entity's recent behavior. Its
knowledge state is $K_{\varnothing}$, so it performs no outcome-supervised
adaptation.

\subsubsection{Reflexion}
\label{app:reflexion-baseline}

Fraud decisions cannot literally reproduce Reflexion's
attempt--feedback--retry loop because each transaction is scored only once. We
therefore map an episode to one week: the agent scores that week's
transactions, waits for their outcomes to mature, writes a verbal reflection
on its mistakes, and carries the resulting memory into subsequent weeks. This
preserves the attempt--feedback--reflection structure across successive
non-retriable batches.

The baseline uses the same weekly cadence and mechanically enforced maturation
wall as SR-Fraud: decisions in cycle $j$ use knowledge derived only from cycles
matured by $j-2$. Its reflection receives the same evidence as SR-Fraud:
confirmed false negatives, the observable risk-3 boundary traffic used for
rejection inference, and the week's highest-dollar confirmed fraud. 
At decision time, the memory is static context in the system prompt, and each
transaction requires one LLM call; memory writing occurs offline once per week
and adds no iterative calls to the live path.

The deliberate contrast variable is \emph{verification}. Reflexion maintains a
single running free-text knowledge document, which the model rewrites weekly to
merge new patterns and remove stale ones. Whatever the model concludes is
placed directly into the following week's prompt. Unlike SR-Fraud, candidate
rules receive no legitimate-traffic rarity screen, fraud-enrichment test, or
promotion gate conditioned on matured precision. We use the least-guarded
reflection prompt, without instructions that approximate these checks, so the
comparison isolates unverified verbal reflection rather than a partially
self-verified variant.

\subsection{Model Configuration}
\label{app:model-config}
We use \texttt{claude-opus-4-5} \cite{anthropic2025opus45}, 
\texttt{claude-sonnet-4-6} \cite{anthropic2026sonnet46},
 and \texttt{gpt-5.4} \cite{openai2026gpt54} through an OpenAI-compatible
gateway. Single-pass scoring uses no extended thinking,
temperature $0.3$ where supported, and default top-$p$. We extract the first
\texttt{RISK\_LEVEL: <1-5>} match; an unparseable response is treated
as allow.

\section{Additional Experiments}
\label{app:additional-analyses}

The supplementary experiments are organized into mechanism checks, verifier
robustness analyses, and deployment-efficiency comparisons.

\subsection{Statistical Reliability}
\label{app:bootstrap}\label{app:bootstrap-processed}

We quantify the reliability of SR-Fraud's improvements with an entity-clustered
bootstrap ($B{=}2000$). Entities are resampled with replacement, and the same
resampled entities are used for both systems so that each paired difference
accounts for the shared sample. Table~\ref{tab:bootstrap-delta} reports the
SR-Fraud-minus-baseline differences and their $95\%$ CIs by prior-history cohort;
bold entries denote intervals that exclude zero.

\begin{table}[t]
\centering
\small
\setlength{\tabcolsep}{2pt}
\begin{tabular}{lcccc}
\toprule
 & \textbf{Prec.} & \textbf{Rec.} & \textbf{F1} & \textbf{Recall$_{\$}$} \\
\midrule
\multicolumn{5}{l}{\emph{$\geq\!0$ prior (all)}}\\
$\Delta$ vs Dec.\ agent & \makecell{\textbf{+5.4}\\{\scriptsize[2.8,8.1]}} & \makecell{\textbf{+7.3}\\{\scriptsize[4.0,10.7]}} & \makecell{\textbf{+6.2}\\{\scriptsize[3.6,9.0]}} & \makecell{\textbf{+19.3}\\{\scriptsize[14.3,24.5]}} \\
$\Delta$ vs CatB.\ period. & \makecell{+0.3\\{\scriptsize[-4.7,5.1]}} & \makecell{\textbf{+8.1}\\{\scriptsize[1.7,14.4]}} & \makecell{+3.8\\{\scriptsize[-1.5,8.8]}} & \makecell{\textbf{+7.8}\\{\scriptsize[0.6,15.2]}} \\
\midrule
\multicolumn{5}{l}{\emph{$\geq\!1$ prior}}\\
$\Delta$ vs Dec.\ agent & \makecell{+2.0\\{\scriptsize[-1.7,5.4]}} & \makecell{+3.5\\{\scriptsize[-0.2,7.4]}} & \makecell{+2.7\\{\scriptsize[-0.6,5.9]}} & \makecell{\textbf{+12.4}\\{\scriptsize[7.0,18.3]}} \\
$\Delta$ vs CatB.\ period. & \makecell{+3.7\\{\scriptsize[-3.0,10.3]}} & \makecell{\textbf{+16.9}\\{\scriptsize[9.0,24.7]}} & \makecell{\textbf{+10.4}\\{\scriptsize[3.3,17.2]}} & \makecell{\textbf{+14.4}\\{\scriptsize[5.1,24.1]}} \\
\midrule
\multicolumn{5}{l}{\emph{$\geq\!2$ prior}}\\
$\Delta$ vs Dec.\ agent & \makecell{0.0\\{\scriptsize[-4.4,4.2]}} & \makecell{+0.6\\{\scriptsize[-3.8,5.2]}} & \makecell{+0.2\\{\scriptsize[-3.7,4.1]}} & \makecell{\textbf{+7.9}\\{\scriptsize[0.3,15.3]}} \\
$\Delta$ vs CatB.\ period. & \makecell{+6.4\\{\scriptsize[-2.1,15.5]}} & \makecell{\textbf{+24.0}\\{\scriptsize[13.9,34.6]}} & \makecell{\textbf{+15.7}\\{\scriptsize[7.1,25.1]}} & \makecell{\textbf{+23.6}\\{\scriptsize[11.7,36.0]}} \\
\midrule
\multicolumn{5}{l}{\emph{$\geq\!3$ prior}}\\
$\Delta$ vs Dec.\ agent & \makecell{+0.6\\{\scriptsize[-5.1,6.4]}} & \makecell{+1.8\\{\scriptsize[-4.0,7.3]}} & \makecell{+1.1\\{\scriptsize[-4.0,6.0]}} & \makecell{+10.4\\{\scriptsize[-0.4,20.8]}} \\
$\Delta$ vs CatB.\ period. & \makecell{+9.9\\{\scriptsize[-1.2,20.9]}} & \makecell{\textbf{+26.5}\\{\scriptsize[14.6,38.4]}} & \makecell{\textbf{+18.9}\\{\scriptsize[8.6,29.3]}} & \makecell{\textbf{+29.9}\\{\scriptsize[14.5,44.8]}} \\
\bottomrule
\end{tabular}
\caption{Paired SR-Fraud-minus-baseline differences with entity-clustered
bootstrap $95\%$ confidence intervals. Bold entries exclude zero.}
\label{tab:bootstrap}\label{tab:bootstrap-delta}
\end{table}

The reflection gain over the decision agent is reliable on the full benchmark:
all four differences are positive with CIs above zero, including $+6.2$ F1
$[3.6,9.0]$ and $+19.3$ Recall$_{\$}$ $[14.3,24.5]$. Within smaller history
cohorts, the precision/recall/F1 intervals include zero. The dollar-recall gain
excludes zero for $\geq1$ ($+12.4$ $[7.0,18.3]$) and $\geq2$ ($+7.9$
$[0.3,15.3]$), while the $\geq3$ interval includes zero ($+10.4$
$[-0.4,20.8]$).

The comparison with CatBoost-periodical follows a different pattern. The methods
have indistinguishable precision and F1 on the full benchmark, while SR-Fraud's
recall and dollar recall are significantly higher. From the $\geq1$ cohort
onward, recall, F1, and dollar-recall intervals exclude zero and widen with
history depth, reaching $+29.9$ Recall$_{\$}$ $[14.5,44.8]$ at $\geq3$ prior
transactions. Thus, reflection provides a robust overall gain over the frozen
decision agent, while the LLM's advantage over the tabular model concentrates
where behavioral history is deepest (\S\ref{sec:context-depth}).

\subsection{History-Content Utilization}
\label{app:history-shuffle}

Section~\ref{sec:context-depth} shows that history \emph{presence} helps. We
further test whether the decision agent uses the \emph{content} of the Hybrid
Episodic Window, or whether any plausible-looking history would do. On the
$\geq\!3$-prior-history cohort ($5{,}650$ transactions, $223$ fraud, $1{,}771$
entities), we score the identical frozen agent (Sonnet~4.6, block at
risk~$\geq\!4$) on the identical current transactions under three history
conditions: \textbf{correct} (the transaction's real window), \textbf{shuffled}
(the recent history of a different, randomly chosen entity, with the
current-transaction fields left unchanged; verified $0$ self-donors), and
\textbf{no-history} (current transaction only). Only the history block changes.

\begin{table}[h]
\centering
\footnotesize
\setlength{\tabcolsep}{2pt}
\begin{tabular}{lccccc}
\toprule
\textbf{History} & \textbf{$\Delta$P} & \textbf{$\Delta$R} &
\textbf{$\Delta$F1} & \textbf{$\Delta$R$_{\$}$} & \textbf{Block} \\
\midrule
CatB.-period. (ref.) & 0.0 & 0.0 & 0.0 & 0.0 & N/A \\
Correct & +9.3 & +24.7 & +17.8 & +19.5 & 4.50\% \\
Shuffled & -24.9 & +53.8 & -14.7 & +51.6 & 60.6\% \\
No-history & -9.1 & -9.8 & -10.3 & -11.8 & 1.89\% \\
\bottomrule
\end{tabular}
\caption{History-content ablation as percentage-point differences from CatBoost-periodical on the $\geq3$ history cohort.}
\label{tab:history-shuffle}
\end{table}

The history-content result is also visible relative to CatBoost-periodical.
Correct history improves precision, recall, F1, and dollar-weighted recall by
$+9.3$, $+24.7$, $+17.8$, and $+19.5$ points. Wrong-entity history produces a
large negative precision delta and a $60.6\%$ block rate, whereas removing
history yields negative recall and dollar-recall deltas. The same pattern holds
across three donor seeds and with amount-matched donors, confirming that the
agent uses entity-specific behavior rather than a generic history indicator.

\subsection{Proposer Ablation}
\label{app:proposer-ablation}

We ask whether a classical rule-discovery algorithm can replace the LLM in the
weekly proposal stage. All runs use the same decision agent (Sonnet~4.6),
maturation wall, verifier, admission budget, and rejection-inference release.
Each rule miner independently constructs its pre-replay state and weekly updates;
none is initialized from LLM-authored rules or state.

\paragraph{Brute-force search is infeasible on raw features.}
Direct search over the raw feature space faces two practical obstacles
(Table~\ref{tab:raw-proposer}). First, discretization produces $527$ atoms, making
an arity-$3$ enumeration $2.4\times10^{7}$ candidates per cycle, nearly
$3{,}000\times$ the $8{,}473$ candidates in the distilled grammar. Every
candidate would also need weekly re-verification on matured evidence. Second, a
tractable beam approximation over-specializes: greedily extending the
highest-precision conjunctions produces extremely narrow rules.
The beam policy attains high marginal precision on the transactions it blocks, but covers under 0.1\% of benchmark traffic and correspondingly loses 31.0 F1 points and 48.2 dollar-recall points relative to static CatBoost. Greedy precision maximization therefore converges on rules that are pure but operationally inert.
The resulting policy fires on just $15$ transactions, yielding high purity but
near-zero coverage.

\paragraph{Comparison with established rule miners.}
We next evaluate three established non-LLM miners as drop-in proposers:
\textbf{FP-Growth} \cite{han2000mining}, which mines frequent itemsets among fraud examples and ranks
them by lift; \textbf{RIPPER} \cite{cohen1995fast}, an imbalance-aware sequential-covering method; and
\textbf{WRAcc subgroup discovery} \cite{han2000mining}, which performs beam search using weighted
relative accuracy. Each method constructs its own initial state from the training
and validation splits and proposes weekly updates through the shared harness. We
evaluate each miner on both the reflection agent's selected feature subset and the
full raw feature set. Two simple controls provide lower bounds: a random-vocabulary
initialization built on the training and validation splits and then held fixed
during replay, and CART induction, which finds no fraud-enriched leaf under the
$0.6\%$ training prevalence even at depth~$8$.
Table~\ref{tab:proposer-ablation} reports the resulting policies.

\begin{table*}[h]
\centering
\small
\setlength{\tabcolsep}{4pt}
\begin{tabular}{lcc}
\toprule
& \textbf{Distilled} & \textbf{Raw} \\
& \textbf{grammar} & \textbf{features} \\
\midrule
Atom pool size & 37 & 527 \\
Arity-3 candidates / cycle & 8{,}473 & $2.4\!\times\!10^{7}$ \\
\bottomrule
\end{tabular}
\caption{The raw feature space a deterministic proposer faces without the LLM.
Exhaustive enumeration over the raw atoms is intractable ($2.4\times10^{7}$
candidates/cycle).}
\label{tab:raw-proposer}
\end{table*}

\begin{table*}[htbp]
\centering
\small
\setlength{\tabcolsep}{3.5pt}
\begin{tabular}{lccccc}
\toprule
\textbf{Proposer (full loop, Sonnet dec.)} & \textbf{$\Delta$P} & \textbf{$\Delta$R} & \textbf{$\Delta$F1} & \textbf{$\Delta$R$_{\$}$} & \textbf{Block} \\
\midrule
CatBoost (reference) & 0.0 & 0.0 & 0.0 & 0.0 & 5.00\% \\
No reflection ($K_{\varnothing}$) & -0.9 & +5.0 & +1.8 & -7.6 & 5.97\% \\
\midrule
Random vocabulary (frozen) & -2.3 & +7.9 & +2.0 & -1.5 & 6.75\% \\
Decision-tree induction & \multicolumn{4}{c}{\emph{no fraud-enriched leaf found}} & n/a \\
\midrule
\multicolumn{6}{l}{\emph{Rule miners over the reflection agent's features}}\\
\quad FP-Growth & +1.9 & +8.2 & +4.8 & +4.3 & 5.94\% \\
\quad RIPPER & +1.1 & +8.1 & +4.3 & +2.1 & 6.06\% \\
\quad WRAcc subgroup & +0.8 & +11.4 & +5.4 & +9.6 & 6.64\% \\
\midrule
\multicolumn{6}{l}{\emph{Rule miners over raw features}}\\
\quad FP-Growth & +1.9 & +5.4 & +3.6 & -2.3 & 5.52\% \\
\quad RIPPER & +0.8 & +7.2 & +3.8 & +0.9 & 5.99\% \\
\quad WRAcc subgroup & +4.8 & +0.9 & +2.7 & -11.1 & 4.47\% \\
\midrule
\textbf{SR-Fraud} (LLM proposer) & +4.5 & +12.3 & +8.0 & +11.7 & 6.08\% \\
\bottomrule
\end{tabular}
\caption{Proposer comparison as percentage-point differences from static
CatBoost. Each rule miner constructs its own initial state and weekly proposals;
the decision agent, maturation wall, verifier, admission budget, and release
mechanism are shared. The independently built random-vocabulary control is held
fixed during replay. Each miner is evaluated on the agent-selected and raw
feature sets. Operating points are not friction-matched.}
\label{tab:proposer-ablation}
\end{table*}

Two patterns emerge. First, SR-Fraud yields the strongest F1 and dollar-recall
point estimates among the evaluated proposers. It reaches
an F1 delta of $+8.0$ and a dollar-recall delta of $+11.7$ points, versus at most
$+5.4$ and $+9.6$ points, respectively,
across the six classical-miner configurations. The simple controls provide little
additional evidence: the frozen random initialization changes F1 by only $0.2$
points relative to no reflection, and CART finds no usable leaf.

Second, the reflection agent's selected feature subset improves both F1 and
dollar recall for every classical miner. For FP-Growth, RIPPER, and WRAcc,
respectively,
the selected/raw F1 deltas are $+4.8$/$+3.6$, $+4.3$/$+3.8$, and
$+5.4$/$+2.7$ points; the corresponding dollar-recall deltas are
$+4.3$/$-2.3$, $+2.1$/$+0.9$, and $+9.6$/$-11.1$ points.
The raw-feature WRAcc run is especially weak on dollar recall because its broad
subgroups fail the shared evidence gates. Thus, adding raw features does not close
the gap. Together, these results support two complementary roles for the LLM:
discovering a compact feature subset and proposing transferable rules within it.

\subsection{Reflexion Operating-Point Sensitivity}
\label{app:reflexion-threshold}

Because the decision agent emits a discrete risk level in $\{1,\dots,5\}$ rather
than a continuous score, the Reflexion baseline cannot be thresholded to an exact
$5\%$ block budget; only two natural cut points exist: risk~$\geq\!4$ (SR-Fraud's
block rule) and the most conservative risk~$\geq\!5$. Table~\ref{tab:reflexion-threshold}
reports both, for all three backbones, under the identical weekly loop and
maturation wall of \S\ref{app:reflexion-baseline} (each backbone serves as both
the decision agent and the reflection author).

\begin{table*}[h]
\centering
\small
\begin{tabular}{llccccc}
\toprule
\textbf{Backbone} & \textbf{Cut} & \textbf{$\Delta$Prec.} & \textbf{$\Delta$Rec.} & \textbf{$\Delta$F1} & \textbf{$\Delta$Recall$_{\$}$} & \textbf{Block} \\
\midrule
CatBoost (reference) & 5\% & 0.0 & 0.0 & 0.0 & 0.0 & 5.00\% \\
Sonnet~4.6 & $\geq\!4$ & -13.3 & +23.5 & -3.9 & +26.1 & 14.80\% \\
Sonnet~4.6 & $\geq\!5$ & +7.0 & +1.8 & +4.2 & -6.6 & 4.34\% \\
Opus~4.5 & $\geq\!4$ & -22.2 & +31.7 & -14.8 & +19.5 & 32.24\% \\
Opus~4.5 & $\geq\!5$ & -10.7 & -15.8 & -13.6 & -30.0 & 3.79\% \\
GPT-5.4 & $\geq\!4$ & -25.1 & +25.4 & -19.6 & +10.8 & 41.44\% \\
GPT-5.4 & $\geq\!5$ & -22.7 & -9.5 & -18.8 & -21.3 & 12.06\% \\
\midrule
\textbf{SR-Fraud} (Sonnet~4.6) & $\geq\!4$ & \textbf{+4.5} & \textbf{+12.3} & \textbf{+8.0} & \textbf{+11.7} & 6.08\% \\
\bottomrule
\end{tabular}
\caption{Reflexion operating points as percentage-point differences from static CatBoost.}
\label{tab:reflexion-threshold}
\end{table*}

At risk~$\geq4$, every Reflexion backbone exceeds the target block budget,
with realized block rates ranging from $14.8\%$ to $41.4\%$. Raising the
threshold to risk~$\geq5$ does not provide a consistent remedy: Sonnet moves
near budget, Opus becomes overly conservative, and GPT-5.4 still blocks well
above budget. The free-text memory therefore inherits each backbone's raw
escalation tendency. SR-Fraud instead bounds policy changes through
matured-evidence admission and a fixed per-cycle cap. Full SR-Fraud is
evaluated with Sonnet~4.6 only, so the cross-backbone comparison is used to
diagnose Reflexion rather than to claim measured SR-Fraud performance on every
backbone.

\subsection{CatBoost at SR-Fraud's Block Budget}
\label{app:reverse-budget-match}

The ordinal LLM policy does not provide a principled within-tier ranking for
forcing it to an arbitrary $5\%$ budget. We therefore leave SR-Fraud completely
unchanged and instead evaluate the continuous CatBoost scores at SR-Fraud's
native operating point \ref{tab:reverse-budget-match}. Full SR-Fraud blocks $982$ of $16{,}140$ transactions
($6.0843\%$, reported as $6.08\%$). For each CatBoost variant, we rank transactions by its continuous
score and block exactly the top $982$; labels are used only after selection to
compute metrics. No SR-Fraud decision or evaluation row is removed.

\begin{table}[h]
\centering
\footnotesize
\setlength{\tabcolsep}{2pt}
\begin{tabular}{lccccc}
\toprule
\textbf{Method} & \textbf{$\Delta$P} & \textbf{$\Delta$R} &
\textbf{$\Delta$F1} & \textbf{$\Delta$R$_{\$}$} & \textbf{Block} \\
\midrule
CatBoost (reference) & 0.0 & 0.0 & 0.0 & 0.0 & 6.08\% \\
CatBoost-periodical & +3.8 & +4.6 & +4.1 & +5.4 & 6.08\% \\
\textbf{SR-Fraud} & \textbf{+6.8} & \textbf{+8.2} &
\textbf{+7.5} & \textbf{+7.3} & 6.08\% \\
\bottomrule
\end{tabular}
\caption{Performance at a common block count as percentage-point differences
from the budget-matched CatBoost reference.}
\label{tab:reverse-budget-match}
\end{table}

\begin{table}[h]
\centering
\footnotesize
\setlength{\tabcolsep}{2pt}
\begin{tabular}{lcccc}
\toprule
\textbf{Method} & \multicolumn{2}{c}{\textbf{Week-8 burst}} &
\multicolumn{2}{c}{\textbf{High-value vector}} \\
 & \textbf{$\Delta$R} & \textbf{$\Delta$R$_{\$}$} &
\textbf{$\Delta$R} & \textbf{$\Delta$R$_{\$}$} \\
\midrule
CatBoost (reference) & 0.0 & 0.0 & 0.0 & 0.0 \\
CatBoost-periodical & +10.2 & +6.3 & -1.8 & -0.2 \\
\textbf{SR-Fraud} & \textbf{+34.5} & \textbf{+35.3} &
\textbf{+9.1} & \textbf{+8.7} \\
\bottomrule
\end{tabular}
\caption{Subset performance at the common $6.08\%$ block count as
percentage-point differences from CatBoost.}
\label{tab:reverse-budget-slices}
\end{table}

Table~\ref{tab:reverse-budget-slices} shows the same point-estimate ordering on
the burst and high-value subsets under the exact matched budget.

At the common block count, SR-Fraud has higher point estimates than both tree
baselines on all four metrics. An entity-clustered paired bootstrap
($B=2000$) gives SR-Fraud-minus-static-CatBoost differences of $+6.8$ precision
$[2.1,11.5]$, $+8.2$ recall $[1.3,14.6]$, $+7.5$ F1 $[2.1,12.7]$, and $+7.3$
dollar recall $[-1.1,15.2]$. Against CatBoost-periodical, the differences are
$+3.1$ precision $[-1.8,7.6]$, $+3.7$ recall $[-2.8,10.4]$, $+3.3$ F1
$[-1.8,8.4]$, and $+1.9$ dollar recall $[-5.4,9.1]$. Thus, after exact block-count
matching, SR-Fraud's precision, recall, and F1 advantages over static CatBoost are
significant (dollar recall is positive but its interval includes zero), while none
of its differences from CatBoost-periodical is statistically significant.

\subsection{Reflection- and Decision-Backbone Combinations}
\label{app:backbone-matrix}

The main results pair each decision agent with its own model as the reflection
agent (the diagonal, $R\!=\!D$). Here we vary the two roles independently: the
reflection agent $R$ (which proposes boundary hypotheses each cycle) and the
decision agent $D$ (which produces the frozen risk scores). All cells use the same
deterministic verifier, rejection-inference release, and per-cycle admission cap.

\begin{table}[h]
\centering
\small
\setlength{\tabcolsep}{4pt}
\begin{tabular}{ll cccc}
\toprule
\textbf{Reflect.} $R$ & \textbf{Decision} $D$ & \textbf{$\Delta$P} & \textbf{$\Delta$R} & \textbf{$\Delta$F1} & \textbf{$\Delta$R$_{\$}$} \\
\midrule
\multirow{3}{*}{Sonnet~4.6}
 & Sonnet~4.6 & +4.5 & +12.3 & +8.0 & +11.7 \\
 & Opus~4.5   & +6.4 & +6.0 & +6.2 & +1.2 \\
 & GPT-5.4    & -1.0 & +6.3 & +2.3 & +1.4 \\
\midrule
\multirow{3}{*}{Opus~4.5}
 & Sonnet~4.6 & +5.3 & +12.7 & +8.7 & +13.2 \\
 & Opus~4.5   & +6.2 & +3.0 & +4.6 & -2.2 \\
 & GPT-5.4    & -1.3 & +3.4 & +0.9 & -3.8 \\
\midrule
\multirow{3}{*}{GPT-5.4}
 & Sonnet~4.6 & +4.9 & +11.8 & +8.1 & +8.7 \\
 & Opus~4.5   & +5.9 & +5.2 & +5.5 & +0.8 \\
 & GPT-5.4    & -1.2 & +7.6 & +2.7 & +2.8 \\
\bottomrule
\end{tabular}
\caption{Reflection ($R$) $\times$ decision ($D$) backbone matrix as
percentage-point differences from static CatBoost. Each cell runs the same
propose--verify loop and rejection-inference release. The main table reports the
diagonal ($R\!=\!D$).}
\label{tab:backbone-matrix}
\end{table}

Two patterns emerge. The \emph{decision} agent sets the operating point (columns):
$D=\,$Sonnet yields the strongest F1/dollar-recall for every reflector, because it
routes the most fraud into the reachable $r{=}3$ band, whereas $D=\,$Opus and
$D=\,$GPT leave a larger share at $r\!\le\!2$ (out of reflection's reach) and are
correspondingly capped. The \emph{reflection} agent matters much less within a
column: with $D=\,$Sonnet,
the F1 deltas are $+8.0$/$+8.7$/$+8.1$ points for $R\in\{$Sonnet, Opus,
GPT$\}$,
indicating that the propose--verify--refine loop is robust to the proposer
backbone.

\subsection{Rejection-Inference Validation}
\label{app:rejection-validation}

Across all replay cycles, the rejection-inference proxy and oracle labels
select the same anonymized release family, with the proxy choosing a slightly
broader band than the oracle.
Relative to static CatBoost, the proxy configuration has recall and dollar-recall
deltas of $+12.3$ and $+11.7$ points; with oracle labels, they become $+15.0$
and $+14.2$ points. The block rate changes from $6.08\%$ to $6.55\%$, and the
selected release family never changes.
Because release is restricted to the
matching risk-4 slice, the resulting policy difference remains localized. This
agreement supports the risk-3 bank as a practical proxy, while not implying that
it identifies the unobserved counterfactual labels of individual blocked
transactions.

\subsection{Verification Confidence Sensitivity}
\label{app:z-sensitivity}

The admission and release bounds use a one-sided Wilson $z=1.0$. We re-verify the persisted main-run
candidates under $z\in\{1.64,1.96,2.58\}$ while holding the remaining pipeline
fixed; $z{=}1.0$ is the reported run. The additional values are common
standard-normal critical values spanning approximately $95\%$ to $99.5\%$
one-sided coverage; they serve as sensitivity settings rather than family-wise
confidence guarantees. Table~\ref{tab:z-sensitivity} shows that the operating
point remains stable across these settings.

\begin{table}[h]
\centering
\footnotesize
\setlength{\tabcolsep}{2pt}
\begin{tabular}{lcccccc}
\toprule
\textbf{Setting} & \textbf{Final rules} & \textbf{$\Delta$P} & \textbf{$\Delta$R} & \textbf{$\Delta$F1} & \textbf{$\Delta$R$_{\$}$} & \textbf{Block} \\
\midrule
CatB. (ref.) & -- & 0.0 & 0.0 & 0.0 & 0.0 & 5.00\% \\
$z=1.0$ & 11 & +4.5 & +12.3 & +8.0 & +11.7 & 6.08\% \\
$z=1.64$ & 6 & +3.8 & +10.9 & +7.1 & +8.0 & 6.00\% \\
$z=1.96$ & 6 & +3.8 & +10.9 & +7.1 & +8.0 & 6.00\% \\
$z=2.58$ & 6 & +4.2 & +9.7 & +6.8 & +6.6 & 5.76\% \\
\bottomrule
\end{tabular}
\caption{Wilson-bound sensitivity as percentage-point differences from static
CatBoost. Candidates are re-verified with only $z$ changed.}
\label{tab:z-sensitivity}
\end{table}

\subsection{Candidate Admissions and Multiplicity}
\label{app:multiplicity}

Table~\ref{tab:candidate-funnel} reports the proposal--verification funnel for
the headline Sonnet configuration. The persisted atom counts are the unions
across iterative proposal rounds. Candidate checks count verifier evaluations
across rounds and therefore need not be unique rules. A verified candidate clears
the evidence gate; cap-excluded candidates clear that gate but fall outside the
fixed per-cycle admission budget.

\begin{table*}[h]
\centering
\small
\setlength{\tabcolsep}{3.5pt}
\resizebox{\columnwidth}{!}{%
\begin{tabular}{ccccccc}
\toprule
\textbf{Cycle} & \textbf{Atoms A/Q/R} & \textbf{Checks} &
\textbf{Verified} & \textbf{Cap-excl.} & \textbf{Admitted} & \textbf{Retired} \\
\midrule
3 & 8/33/3  & 1{,}092 & 0 & 0 & 0 & 0 \\
4 & 10/34/5 & 1{,}449 & 0 & 0 & 0 & 0 \\
5 & 10/34/5 & 1{,}612 & 7 & 2 & 5 & 0 \\
6 & 10/36/5 & 1{,}378 & 0 & 0 & 0 & 2 \\
7 & 11/36/5 & 1{,}467 & 1 & 0 & 1 & 0 \\
8 & 11/37/4 & 1{,}440 & 0 & 0 & 0 & 0 \\
9 & 11/37/5 & 1{,}351 & 2 & 0 & 2 & 0 \\
\midrule
\textbf{Total} & -- & \textbf{9{,}789} & \textbf{10} & \textbf{2} & \textbf{8} & \textbf{2} \\
\bottomrule
\end{tabular}
}
\caption{Headline proposal--verification funnel by replay cycle. A/Q/R denotes
amount atoms, qualifier atoms, and release bands. Checks are verifier evaluations
across iterative rounds, not deduplicated candidate rules.}
\label{tab:candidate-funnel}
\end{table*}

Table~\ref{tab:funnel} traces the active-signature count per cycle for each
self-reflection configuration ($R\!=\!D$), together with the total admissions and
retirements. Because $r{=}3$ evidence only matures after the two-week wall, no
signature can be corroborated before cycle~4; Opus and GPT first grow at cycle~4,
whereas Sonnet first admits new signatures at cycle~5. Each admits its verified
families over the following cycles.
The three backbones differ sharply in admission behavior: GPT admits the maximum
$5$ signatures at every growth cycle ($20$ total, cap-bound), Opus admits $11$
without ever retiring, and Sonnet is the most conservative ($8$ admissions) and the
only configuration to retire ($-2$ at cycle~6, so its count dips from $10$ to $8$
before regrowing to $11$). Per-cycle corroboration and the fixed cap limit growth
but are not a formal multiple-testing correction. The \emph{aggregate} effect of
these online admissions is quantified in Table~\ref{tab:static-state}: relative
to the evolved system, holding the pre-replay state fixed reduces F1 by
$0.4$/$2.4$/$1.7$ points and dollar-recall by $4.3$/$11.2$/$5.9$ points for
Sonnet/Opus/GPT.

\begin{table*}[h]
\centering
\small
\setlength{\tabcolsep}{3.5pt}
\begin{tabular}{lccccccc c cc}
\toprule
\textbf{Active sigs.\ after cycle} & 1 & 3 & 4 & 5 & 6 & 7 & 9 & & $\Sigma$adm. & $\Sigma$ret. \\
\midrule
Sonnet$\times$Sonnet & 5 & 5 & 5 & 10 & 8 & 9 & 11 & & 8 & 2 \\
Opus$\times$Opus     & 9 & 9 & 14 & 19 & 19 & 19 & 20 & & 11 & 0 \\
GPT$\times$GPT       & 5 & 5 & 10 & 10 & 15 & 20 & 25 & & 20 & 0 \\
\bottomrule
\end{tabular}
\caption{Active escalation-signature count per reflection cycle for each
self-reflection ($R\!=\!D$) configuration, from the state deployed at the start
of replay through the state after the final cycle, with total admissions
($\Sigma$adm.) and retirements ($\Sigma$ret.). Growth is gated by per-cycle corroboration on matured evidence and a fixed
admission cap of $5$ per cycle. GPT saturates that cap at every growth cycle
($+5$ each), indicating it proposes more corroborated candidates than the cap admits;
Sonnet is the most selective and the only configuration to retire rules ($-2$ at
cycle~6 on later evidence).}
\label{tab:funnel}
\end{table*}

\subsection{Online Reflection vs.\ a Static State}
\label{app:static-state}

To isolate the contribution of \emph{online} reflection, we compare each evolved
configuration against a control that holds fixed the pre-replay knowledge state
constructed from the training and validation splits. The decision agent, initialization procedure,
verifier, and rejection-inference release are otherwise unchanged.
Table~\ref{tab:static-state} reports both. Online reflection improves
dollar-recall for every backbone, especially where the static state is weakest
relative to the decision agent's reachable pool, confirming that the per-cycle
propose--verify updates---rather than a one-time offline construction---drive the
gains.

\begin{table}[h]
\centering
\small
\setlength{\tabcolsep}{4pt}
\resizebox{\columnwidth}{!}{%
\begin{tabular}{llcccc}
\toprule
\textbf{Config} ($R\!=\!D$) & \textbf{State} & \textbf{$\Delta$P} & \textbf{$\Delta$R} & \textbf{$\Delta$F1} & \textbf{$\Delta$R$_{\$}$} \\
\midrule
\multirow{2}{*}{Sonnet~4.6} & Static state (reference) & 0.0 & 0.0 & 0.0 & 0.0 \\
                            & Evolved state (SR-Fraud) & +0.4 & +0.5 & +0.4 & +4.3 \\
\midrule
\multirow{2}{*}{Opus~4.5}   & Static state (reference) & 0.0 & 0.0 & 0.0 & 0.0 \\
                            & Evolved state (SR-Fraud) & 0.0 & +4.2 & +2.4 & +11.2 \\
\midrule
\multirow{2}{*}{GPT-5.4}    & Static state (reference) & 0.0 & 0.0 & 0.0 & 0.0 \\
                            & Evolved state (SR-Fraud) & +1.4 & +2.1 & +1.7 & +5.9 \\
\bottomrule
\end{tabular}
}
\caption{Online reflection relative to a static-state control for each
self-reflection configuration. Values are within-backbone percentage-point
differences; the decision agent, initialization procedure, verifier, and release
mechanism are matched.}
\label{tab:static-state}
\end{table}

\subsection{Request-Time Latency Analysis}
\label{app:latency-analysis}

SR-Fraud keeps reflection off the request path. Each decision uses one stateless
LLM call, after which the knowledge state $K$ is applied deterministically.
Table~\ref{tab:latency} compares this single-pass design with Reflexion and an
inference-time reasoning agent, using Sonnet~4.6 and the same Hybrid Episodic
Window. Single-pass SR-Fraud and Reflexion remain near $1.3\,\mathrm{s}$ at
$p_{50}$, whereas chain-of-thought raises latency to $47.3\,\mathrm{s}$ and
greatly increases output length. SR-Fraud instead amortizes reasoning into the
offline reflection cycle, improving the policy without adding request-time
reasoning calls; its distinction from Reflexion is verification quality rather
than latency.

\begin{table}[h]
\centering
\small
\setlength{\tabcolsep}{4pt}
\begin{tabular}{lccc}
\toprule
\textbf{Configuration} & \textbf{$p_{50}$ (s)} & \textbf{$p_{95}$ (s)} & \textbf{Out tok} \\
\midrule
Single-pass / SR-Fraud & 1.31 & 2.41 & 11 \\
Reflexion (prompt memory) & 1.38 & 3.91 & 11 \\
Reasoning/CoT agent & 47.3 & 61.5 & 2{,}484 \\
\bottomrule
\end{tabular}
\caption{Request-time latency and output length for single-pass, prompt-memory, and chain-of-thought decision agents.}
\label{tab:latency}
\end{table}

\subsection{Post-Training and Deployment Analysis}
\label{app:post-training-details}

\paragraph{Newer closed models do not guarantee better fraud detection.}
Across three model families, the newer closed release fails to improve fraud
detection (Table~\ref{tab:model-generation}). Opus~5 and Sonnet~5 regress on all
four metrics versus Opus~4.5 and Sonnet~4.6; the newer GPT-5.6 (sol) variant raises
precision only by becoming far more conservative,
leaving its F1 and dollar-recall well below GPT-5.4. In every family the newer model
lowers recall, F1, and dollar-recall. This non-monotonic behavior motivates adapting
a self-hosted model whose training and deployment are under system control.

\begin{table}[h]
\centering
\footnotesize
\setlength{\tabcolsep}{3pt}
\begin{tabular}{lcccc}
\toprule
\textbf{Decision-agent} & \textbf{$\Delta$P} & \textbf{$\Delta$R} & \textbf{$\Delta$F1} & \textbf{$\Delta$R$_{\$}$} \\
\midrule
CatBoost (reference) & 0.0 & 0.0 & 0.0 & 0.0 \\
Opus~4.5 & -2.1 & -1.5 & -1.8 & -15.3 \\
Opus~5 & -10.8 & -2.3 & -7.3 & -17.2 \\
\midrule
Sonnet~4.6 & -0.9 & +5.0 & +1.8 & -7.6 \\
Sonnet~5 & -11.9 & -17.2 & -15.0 & -30.5 \\
\midrule
GPT-5.4 & -11.8 & +2.8 & -6.4 & -11.0 \\
GPT-5.6 (sol) & -0.3 & -17.9 & -12.5 & -32.8 \\
\bottomrule
\end{tabular}
\caption{Successive closed-model generations as percentage-point differences from static CatBoost.}
\label{tab:model-generation}
\end{table}

\paragraph{Training data and reward.}
We post-train \texttt{Qwen/Qwen3.5-9B} \cite{qwen35} on prompts rendered in the same transaction
format as evaluation. The training split retains all fraud cases and samples
three non-fraud transactions per fraud case, yielding roughly $4{,}000$ prompts.
The reported \textbf{SR-Fraud-PT} row is
this post-trained model evaluated as a single-pass decision agent at the same
$r\!\ge\!4$ operating point, with no symbolic knowledge state applied; it thus
isolates weight-encoded adaptation and is complementary to, not combined with,
the reflection layer. Post-training is essential: the untrained base 9B scores far
below every other system (Table~\ref{tab:post-training-capacity}), so the gain
reflects the domain adaptation rather than the backbone's zero-shot ability. The reward maps risk levels 1--5 to
signed fraud confidence $(-1,-0.5,0,0.5,1)$ and applies class weights 3 for fraud
and 1 for non-fraud, so constant policies have zero expected class-balanced
reward.

\paragraph{Optimization.}
We trained Qwen3.5-9B using Group Relative Policy Optimization
(GRPO; \citealp{shao2024deepseekmath}) with the SkyRL framework \citep{griggs2025skrylv01}, 
using eight completions per
prompt and Low-Rank Adaptation (LoRA; \citealp{hu2022lora}) on all linear layers
(rank 32, $\alpha=64$, Kaiming initialization). The run uses three epochs, train batch size 64, learning rate $5\times10^{-5}$, and KL coefficient $10^{-3}$. It is then evaluated with vLLM~0.20.2 in
bfloat16.

\paragraph{Detection and adaptability.}
Post-training allows the smaller self-hosted model to reach a detection profile
similar to the commercial SR-Fraud configuration and supports local GPU serving.
Its knowledge, however, is encoded in model weights:
tracking new fraud patterns requires another training and deployment cycle. In
contrast, SR-Fraud can incorporate verified symbolic rules between model
refreshes. The two mechanisms are therefore complementary: post-training
provides a strong base policy, while reflection supplies an auditable adaptation
channel between retraining cycles.

\paragraph{Serving and latency measurement.}
We serve the post-trained 9B decision agent locally with vLLM~0.20.2 in bfloat16
on a single NVIDIA~L40S GPU, exposing an OpenAI-compatible endpoint. We measure
request-time latency on 200 held-out prompts at concurrency~10, timing each call
from submission to the first-line risk verdict. On this setup the self-hosted
agent has a median request latency of $0.74\,\mathrm{s}$ and a $p_{90}$ of $0.94\,\mathrm{s}$
(Table~\ref{tab:post-training-capacity}), below the corresponding $p_{90}$ of the
hosted Sonnet~4.6 decision agent and full SR-Fraud ($2.02\,\mathrm{s}$)
reported in Table~\ref{tab:main-results}, which are billed per token and served
through a hosted API. A self-hosted model
therefore trades per-token cost for provisioned GPU capacity while keeping the
single request-time LLM call well within the authorization-path latency budget.

\paragraph{Cost trade-off.}
The two serving modes have opposite cost structures: a hosted API charges a flat
per-token amount per call, whereas a self-hosted GPU is a fixed daily cost
amortized over the day's calls. At $6{,}000$ input and $11$ output tokens per
call, a call costs $1.82$ cents on Sonnet ($\$3/\$15$ per $1$M in/out tokens) and
$3.03$ cents on Opus ($\$5/\$25$). Serving the 9B agent on one GPU costs
$\$34$/day (L40S reserved), $\$54$/day (L40S on-demand), or $\$81$/day (RTX
Pro~6000 on-demand), amortized over the day's volume. The measured throughput of
$12.96$ req/s at concurrency~$10$ corresponds to approximately $1.1$M calls/day
under continuous utilization, so all volumes considered here fit on one GPU.
Table~\ref{tab:cost-breakeven} reports the
resulting per-call cost as a fraction of the Sonnet and Opus API prices: reserved
self-hosting already undercuts Sonnet at $2{,}500$ calls/day, and by $10{,}000$
calls/day every option costs well under a third of the Sonnet price and under a
fifth of Opus. Beyond cost, self-hosting avoids sending transaction prompts to an
external model API and reduces provider-drift risk.

\begin{table}[t]
\centering
\small
\setlength{\tabcolsep}{4pt}
\resizebox{\columnwidth}{!}{%
\begin{tabular}{lcccc}
\toprule
\textbf{Self-hosted 9B} & \multicolumn{4}{c}{\textbf{Calls / day}} \\
\cmidrule(lr){2-5}
($\times$ Sonnet~/~Opus) & 2{,}500 & 5{,}000 & 7{,}500 & 10{,}000 \\
\midrule
L40S (reserved)     & 0.75~/~0.45 & 0.37~/~0.22 & 0.25~/~0.15 & 0.19~/~0.11 \\
L40S (on-demand)    & 1.19~/~0.71 & 0.59~/~0.36 & 0.40~/~0.24 & 0.30~/~0.18 \\
Pro~6000 (on-demand) & 1.78~/~1.07 & 0.89~/~0.53 & 0.59~/~0.36 & 0.44~/~0.27 \\
\bottomrule
\end{tabular}
}
\caption{Self-hosted 9B per-call cost relative to the Sonnet and Opus
API price ($6{,}000$ input / $11$ output tokens; Sonnet $1.82$ cents, Opus $3.03$ cents per
call), as a multiple of each API's cost. Self-hosted cost is the fixed
GPU~$\$$/day divided by daily volume on a single node; values below $1\times$ mean
self-hosting is cheaper.}
\label{tab:cost-breakeven}
\end{table}

\section{Additional Discussion}
\label{sec:discussion}

\textbf{Why additive rather than end-to-end LLM?}
Two considerations argued against replacing the tabular classifier.
\emph{Latency}: the upstream classifier responds in tens of milliseconds;
an LLM round-trip is slower.
\emph{Cost}: at sufficiently high traffic volumes, per-request LLM calls can
dominate compute spend.

\textbf{Why retain human review?}
The reported replay is fully automated. In production, an optional human
checkpoint can review each knowledge-state changelog before deployment. Because
updates are weekly and review occurs offline, this control does not affect
request-time latency.

\end{document}